\documentclass[manuscript]{acmart}
\usepackage{dblfloatfix}
\usepackage{placeins}
\AtBeginDocument{%
  \providecommand\BibTeX{{%
    \normalfont B\kern-0.5em{\scshape i\kern-0.25em b}\kern-0.8em\TeX}}}

\copyrightyear{2020}
\acmYear{2020}
\setcopyright{acmcopyright}
\acmConference[ACSW 2020]{Proceedings of the Australasian Computer Science Week Multiconference}{February 4--6, 2020}{Melbourne, VIC, Australia}
\acmBooktitle{Proceedings of the Australasian Computer Science Week Multiconference (ACSW 2020), February 4--6, 2020, Melbourne, VIC, Australia}
\acmPrice{15.00}
\acmDOI{10.1145/3373017.3373028}
\acmISBN{978-1-4503-7697-6/20/02}

\begin{document}

\title{A Short Survey of Pre-trained Language Models for Conversational AI-A New Age in NLP}


\author{Munazza Zaib}
\affiliation{%
  \institution{Macquarie University}
  \city{Sydney}
  \country{Australia}}
\email{muanzza-zaib.ghori@students.mq.edu.au}

\author{Quan Z. Sheng}
\affiliation{%
  \institution{Macquarie Uiversity}
  \city{Sydney}
  \country{Australia}}
  \email{michael.sheng@mq.edu.au}
  
\author{Wei Emma Zhang}
\affiliation{%
  \institution{The University of Adelaide}
  \city{Adelaide}
  \country{Australia}}
  \email{wei.e.zhang@adelaide.edu.au}

\begin{abstract}
  Building a dialogue system that can communicate naturally with humans is a challenging yet interesting problem of agent-based computing. The rapid growth in this area is usually hindered by the long-standing problem of data scarcity as these systems are expected to learn syntax, grammar, decision making, and reasoning from insufficient amounts of task-specific dataset. The recently introduced pre-trained language models have the potential to address the issue of data scarcity and bring considerable advantages by generating contextualized word embeddings. These models are considered counterpart of ImageNet in NLP and have demonstrated to capture different facets of language such as hierarchical relations, long-term dependency, and sentiment. In this short survey paper, we discuss the recent progress made in the field of pre-trained language models. We also deliberate that how the strengths of these language models can be leveraged in designing more engaging and more eloquent conversational agents. This paper, therefore, intends to establish whether these pre-trained models can overcome the challenges pertinent to dialogue systems, and how their architecture could be exploited in order to overcome these challenges. Open challenges in the field of dialogue systems have also been deliberated. 
\end{abstract}



\maketitle

\section{Introduction}
Despite much progress in the field of natural language processing and intelligent agents, the conversational part for communication between a human and a machine is still in its inception phase. It is only recently that neural generative models had gotten the attention of researchers and improved the field of conversational AI drastically. With a large amount of available `big data' and advanced deep learning methods, the objective of designing digital conversation systems as our virtual assistant is no longer a dream.

Based on functionality, conversational AI can be categorized into three categories: 1) task-oriented systems, 2) chat-oriented systems, and 3) question answering systems \cite{DBLP:conf/sigir/GaoG018}. Task-oriented dialogue systems are designed to complete a specific task on the user's behalf such as booking hotels, making a restaurant reservation or finding products. The second category mainly focuses on carrying out a conversation with the user on open-domain topics, and question answering bots are designed to find an appropriate answer to user's query using all its available knowledge and resources. Though, these systems have come a long way in terms of progress but conversing with such models for even a short amount of time quickly unveils the inconsistency in generated responses.

A different number of strategies have been introduced over a period to address this issue. One of the standard methods of designing an NLP based project is to utilize word embeddings, pre-trained on a huge amount of unlabelled data using distributed word representations such as GloVe and Word2Vec, to initialize the first layer of the neural network. The rest of the layers are then trained on a task-specific dataset. However, these techniques failed to capture the correct context of the word used in a sentence. For example ``an apple a day, keeps the doctor away" and ``I own an Apple Macbook, two ``apple" words refer to very different things but they would still share the same word embedding vector. Recently, the concept of pre-trained language modeling is introduced. The word embeddings generated by these language models are pre-trained on large corpora and are then utilized as either distributed word embeddings or fine-tuned according to the specific task needs. This comes under the category of transfer learning and has achieved state-of-the-art results in various NLP tasks. The introduction of pre-trained language models has given the field of conversational AI a new direction especially to question answering. Thus, we focus more on the question answering systems than the other two dialogue systems as this field has been totally transformed by these pre-trained language models.

In this short survey paper, we first discuss the different pre-trained language modeling techniques introduced till now. Then, we extend our discussion to the implementation of these models in dialogue systems with a special emphasis on question answering systems. In the end, we present the open challenges pertaining to these language models that need to be addressed. 
\section{Pre-trained Language Modeling}
Most datasets available for NLP tasks are rather small. This data scarcity problem makes it difficult to train the deep neural networks, as they would result in an over-fitted model and not generalize well on these small datasets. The concept of pre-training a model on ImageNet corpus has been employed for quite a few years in the field of computer vision \cite{DBLP:conf/nips/PalatucciPHM09,DBLP:conf/aaai/LarochelleEB08}. The idea is to make the model learn the general features of an image and this learning can then be utilized in any vision task such as image captioning etc. to achieve the state-of-the-art results.  

Pre-trained language modeling can be considered an equivalent of ImageNet in NLP and has achieved state-of-the-art results on various downstream NLP tasks such as sentiment analysis, data classification, and question answering, etc. The timeline of pre-trained language models can be divided into two categories:
\begin{figure*}[htb]
    \centering
    \includegraphics[width=\textwidth]{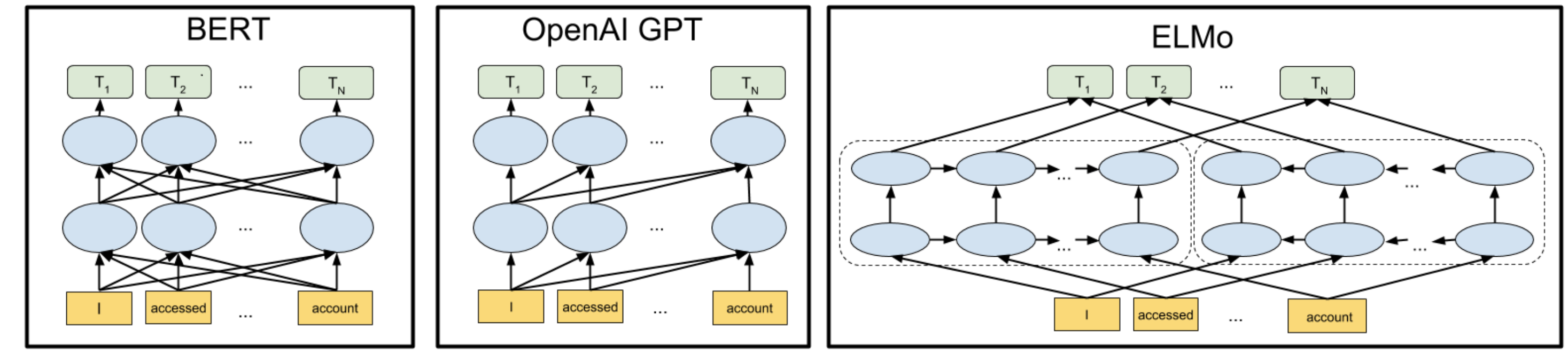}
     \Description[comparison of pre-trained language models]{Architectural comparison of BERT, OpenAI GPT, and ELMo explaining the attention mechanism in them.}
    \caption{Comparison of BERT, OpenAI GPT and ELMo model architectures \cite{DBLP:conf/naacl/DevlinCLT19}.}
    \label{fig:diff}
\end{figure*}

\textbf{Feature based Approaches:} Learning the appropriate representation of the words has been an active research area for a long time. The pre-trained word embeddings provide an edge to modern NLP systems over the embeddings learned from scratch and can be word level, sentence level, or paragraph level based on their granularity. These learned representations are also utilized as features in NLP downstream tasks. 

Embeddings from Language Models (ELMo) \cite{DBLP:conf/naacl/PetersNIGCLZ18} revolutionize the concept of general word embeddings by proposing the idea of extracting the context-sensitive features from the language model. Incorporating these context-sensitive embeddings into task-specific model results in leading-edge results on several NLP tasks such as named entity recognition, sentiments analysis, and question answering on SQuAD \cite{DBLP:conf/emnlp/RajpurkarZLL16} dataset.

\textbf{Fine tuning based Approaches:} There has been a trend of transfer learning from language models recently. The main idea is to pre-train a model on unsupervised corpora and then fine-tune the same model for the supervised downstream task. The advantage of implementing these approaches is that only a few parameters need to be learned from scratch. These models are the adaptation of Google's Transformer model. First in the series is OpenAI's Generative pre-training Transformer (GPT) \cite{radford2018improving}. The model is trained to predict the words only from left-to-right hence, capture the context unidirectionally. The work was further improved by Devlin who introduced a bi-directional pre-trained model called Bi-directional Encoder Representations from Transformers (BERT) \cite{DBLP:conf/naacl/DevlinCLT19}. They addressed the unidirectional constraint by integrating the concept of `Masked Language Model' in their model that randomly masks some of the input tokens and the goal is to predict the masked token based on the captured context from both directions. They also trained their model on `Next Sentence Prediction' task that further improved the model's performance by a good margin. Unlike GPT, BERT is based on Transformer's encoder block which is designed for language understanding. The model achieved state-of-the-art results on 11 NLP downstream tasks including question natural language inference, Quora question pairs, and question answering on SQuAD. Later, Radford et al. \cite{radford2019language} improved their predecessor, GPT, by introducing GPT2. The model is bidirectionally trained on 8M web pages and has 1.5B parameters, 10 times greater than the original model. The model is based on the Transformer's decoder and is designed to generate language naturally.
Although out of all the models BERT seems to perform very well, it still has some loopholes in its implementation. First of all, it uses [MASK] token during the pre-training process, but these token are missing from real data during the fine tuning phase, which results in a pre-train-finetune discrepancy. Another weakness of BERT is that it assumes that masked tokens are independent of each other and are only predicted using unmasked tokens. Recently, another language model, XLNet \cite{DBLP:journals/corr/abs-1906-08237}, improved the shortcomings of BERT by using Transformer-XL as base architecture and introducing the dependency between the masked positions. It is an autoregressive language model that utilizes the context to predict the next word. The context word here is constrained to two directions, either backward or forward. It uses permutation language modeling to overcome the drawbacks of BERT. 

The visual comparison between different language models is shown in Figure~\ref{fig:diff}. It can be seen that BERT is deeply bidirectional, Open AI GPT is unidirectional, and ELMo is shallowly bidirectional.

\section{Pre-trained Language Modeling Approaches for Dialogue Systems}
Although the other two categories are still in infancy in terms of utilizing the pre-trained language models, question answering, also known as machine comprehension (MC), has achieved state-of-the-art results when leveraged the strength of these as their base architecture. 

\subsection{Question Answering System}
Question answering systems come under the field of information retrieval (IR) and natural language processing (NLP), which involves building a system intelligent enough to answer the questions asked by the humans in a natural language. The question answering systems can be categorized as:

\subsubsection{\textbf{Single-turn MC}} There has been rapid a progress after the introduction of pre-trained language models and most of the question answering systems have achieved human-level accuracy on Stanford Question Answering Dataset (SQuAD) \cite{DBLP:conf/emnlp/RajpurkarZLL16}.
The SQuAD dataset is a standard benchmark for the machine comprehension problem consisting of Wikipedia articles and questions posed on those articles by a group of co-workers. The system must select the right answer span for the question from all the possible answers in the given passage.

Word embeddings generated using ELMo are used with BiLM and BiDAF \cite{Seo2016BidirectionalAF} set up to achieve higher accuracy scores on SQuAD leaderboard \footnote{https://rajpurkar.github.io/SQuAD-explorer/}. The results of ELMo are improved by BERT model when used in different architecture settings such as \cite{DBLP:journals/corr/abs-1909-02209} introduced semantic information into the BERT model. Another model introduced by Facebook called RoBERTa \cite{DBLP:journals/corr/abs-1907-11692} improved BERT's performance significantly by training it again on huge corpus and modifying its hyper-parameters more carefully. Recently, the top accuracy score on SQuAD's leader board is held by a model based on XLNet which has managed to improve the accuracy more than the actual human's performance.

\subsubsection{\textbf{Multi-turn MC}}
Multi-turn machine comprehension, also known as conversational machine comprehension (CMC), combines the elements of chit-chat and question answering. The difference between MC and CMC is that questions in CMC form a series of conversations and require the proper modeling of history to comprehend the context of the current question correctly. High-quality conversational datasets such as QuAC \cite{DBLP:conf/emnlp/ChoiHIYYCLZ18} and CoQA \cite{DBLP:journals/tacl/ReddyCM19} have provided the researchers a great source to work deeply in the field of CMC. 

The first BERT based model for QuAC was based on history answer embeddings to provide extra information to input tokens \cite{DBLP:conf/sigir/Qu0QCZI19}. Later, \cite{ohsugi-etal-2019-simple} improved accuracy by introducing the last two contexts when answering the current question. \cite{DBLP:journals/corr/abs-1908-05117} introduced the reasoning process in BERT-based architecture that improved the accuracy on the leader board drastically as compared to the previous models. Currently, the top scores QuAC leaderboard \footnote{http://quac.ai/} are of BERT-based question answering models.  
The BERT and XLNet based models that tested the accuracy on SQuAD dataset also evaluated their models on CoQA. The top positions on CoQA leaderboard \footnote{https://stanfordnlp.github.io/coqa/} are occupied by pre-trained language models. 

\begin{figure}[!h]
    \centering
    \includegraphics[width=\linewidth]{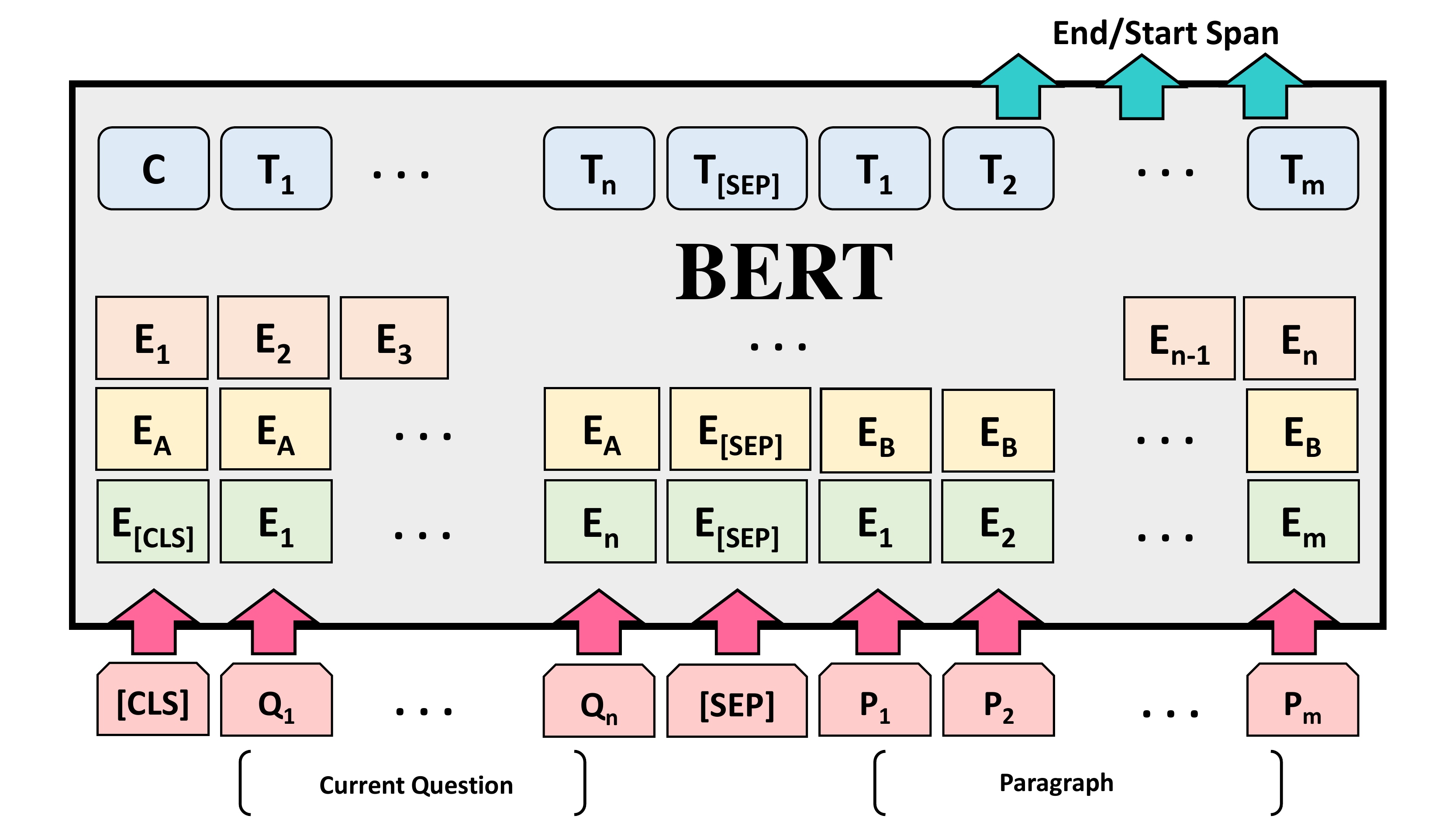}
    \Description[BERT for QA task.]{The adaptation of BERT to suit the task of machine comprehension.}
    \caption{Adaptation of BERT for question answering task}
    \label{fig:arch}
\end{figure}

Figure~\ref{fig:arch} shows how to adapt a BERT-based model for MC or CMC tasks. The input to the model is a question and a paragraph, and the output is the answer span in the given paragraph. A special classification token [CLS] is added before the given question. Then, the question is concatenated with the paragraph into one sequence using [SEP] token. The sequence is provided as an input to the BERT-based model with segment and positional embeddings. Finally, the hidden state of BERT is then converted into the probabilities of start and end answer span by a linear layer and softmax function. 

\subsection{Other Dialogue Systems}
Recent work in large-scale pre-training (such as GPT-2 and BERT) on large text corpus using transformers has attracted significant interests and achieved great empirical success. However, leveraging the strength of massive publicly available colloquial text datasets to build a full-fledge conversational agent (i.e. task-oriented and chat-oriented) is still progressing. The use of these pre-trained language models is in its inception phase and not much work has been done. In this section, we briefly discuss the research carried out in the two areas:
\subsubsection{\textbf{Task-Oriented Dialogue System}}
A traditional task-oriented model consists of four modules namely: i) natural language understanding, ii) dialogue state tracking (DST), iii) policy learning, and iv) natural language generation. The goal of such systems is to assist the user by generating a valuable response. This response generation requires a considerable amount of labeled data fro the training purpose. A question that comes naturally to mind is: Can we take advantage of transfer learning through pre-trained language models to enable the modeling of task-oriented systems. The question has been addressed by \cite{DBLP:journals/corr/abs-1907-05774} which introduces a GPT based framework to evaluate the ability to transfer the generation capability of GPT to task-specific multi-domains. They used multi-domain dataset MultiWoz \cite{DBLP:conf/emnlp/BudzianowskiWTC18} to learn and understand the domain-specific tokens which make it easier to adapt to unseen domains. 
Chao \& Lane \cite{Chao2019} recently utilize the strengths of BERT in improving the scalability of DST module. The DST module is use to maintain the state of user's intentions through out the dialogue. The key component of the model is BERT dialogue context encoding module which generates contextualized representations of the words which are very effective for mining slot values from the contextual patterns. 
 
\subsubsection{\textbf{Chat-Oriented Dialogue System}}
Chat-oriented dialogue systems are known to have several issues such as they are often not very engaging and lack specificity. 
To address these problems, TransferTransfo, a persona-based model, \cite{radford2019language} is introduced. They have extended the transfer learning from language understanding to generative tasks such as open-domain dialogue generation using GPT and addressed the above-mentioned issues by combining many linguistics aspects such as common-sense knowledge, co-reference resolution, and long-range dependency. 
\section{Open Challenges}
Even after the implementation of advanced deep learning techniques and pre-trained language models, it is worth noting that current dialogue systems are far from perfect. Despite the aforementioned progress, there are still some open challenges that need attention. In this section, we highlight those open issues briefly:
\begin{itemize}
    \item \textbf{Privacy:} Dialogue systems entertain and interact with a large number of people. These assistants, with the given ability to learn through interactions, can store user's information that could be sensitive. Hence, it is of utmost importance to keep the user's privacy in mind while designing the system.
    \item \textbf{Empathetic Computing:} Emotion perception and expression are an integral part of designing human-like dialogue systems. The pre-trained language models are known to have achieved state-of-the-art results when it comes to the sentiment analysis of the user's reviews of products and services. But not much work is done to incorporate this concept into dialogue systems to maximize long-term user engagement by generating more thoughtful responses. 
    \item \textbf{Lack of Inference Capability:} Most of the existing systems, particularly question answering systems, are based on semantic matching i.e relevance between the previous context and the current question is calculated to answer the question, which makes the systems incapable of reasoning. This lack of inference capability often results in inaccurate answers. Thus, much effort is required in this area to provide strong reasoning skills to the conversational systems.
\end{itemize}
\section{Conclusion}
In recent years, the promising notion of pre-trained language models has gained widespread attention by researchers. It is an emerging paradigm aimed to generate better contextualized representations of the words so that the dialogue systems have a better understanding of the context. This paper is an effort to investigate the recent trends introduced in language models and their application to the dialogue systems. We focus the discussion on question answering systems as they have been influenced by this progress the most. We, then, briefly highlight the progress in the other two categories. Whilst these pre-trained models have the potential to address most of the limitations posed by previous methods, nevertheless, there are still some open issues that need to be addressed. Thus, in the end, we have highlighted the challenges pertaining to dialogue systems that demand attention.

\bibliographystyle{ACM-Reference-Format}
\bibliography{acmart}

\end{document}